\documentclass[sigconf,authordraft，review]{acmart}
\usepackage{cleveref}
\usepackage{algorithm}
\usepackage{algorithmic}

\AtBeginDocument{%
  }

\setcopyright{acmlicensed}
\copyrightyear{2024}
\acmYear{2024}
\acmDOI{XXXXXXX.XXXXXXX}

\acmConference[CIKM2024]{Make sure to enter the correct conference title from your rights confirmation emai}{October 21–-25,2024}{Boise, Idaho, USA}
  
\acmISBN{978-1-4503-XXXX-X/18/06}

\begin{document}

\title{Towards Efficient Disaster Response via Cost-effective Unbiased Class Rate Estimation through Neyman Allocation Stratified Sampling Active Learning}


\author{Yanbing Bai}
\authornote{Both authors contributed equally to this research.}
\affiliation{%
  \institution{Center for Applied Statistics and School of Statistics, Renmin University of China}
  \city{Beijing}
  \country{China}
}
\email{ybbai@ruc.edu.cn}

\author{Xinyi Wu}
\authornotemark[1]
\affiliation{%
  \institution{Center for Applied Statistics and School of Statistics, Renmin University of China}
  \city{Beijing}
  \country{China}
}
  \email{2021103789@ruc.edu.cn}
  
\author{Lai Xu}
\affiliation{%
 \institution{Independent Researcher}
 \city{Beijing}
 \country{China}}
\email{pkusimon@gmail.com}

\author{Jihan Pei}
\affiliation{%
 \institution{Center for Applied Statistics and School of Statistics, Renmin University of China}
 \city{Beijing}
 \country{China}}
\email{2021201671@ruc.edu.cn}

\author{Erick MAS}
\affiliation{%
 \institution{Disaster Geo-informatics Laboratory, International Research Institute of Disaster Science, Tohoku University}
 \state{Sendai}
 \city{Miyagi}
 \country{Japan}}
\email{mas@irides.tohoku.ac.jp}

\author{Shunichi Koshimura }
\authornote{Corresponding Author.}
\affiliation{%
 \institution{Disaster Geo-informatics Laboratory, International Research Institute of Disaster Science, Tohoku University}
 \state{Sendai}
 \city{Miyagi}
 \country{Japan}}
\email {koshimura@irides.tohoku.ac.jp}


\begin{abstract}
With the rapid development of earth observation technology, we have entered an era of massively available satellite remote-sensing data. However, a large amount of satellite remote sensing data lacks a label or the label cost is too high to hinder the potential of AI technology mining satellite data. Especially in such an emergency response scenario that uses satellite data to evaluate the degree of disaster damage. Disaster damage assessment encountered bottlenecks due to excessive focus on the damage of a certain building in a specific geographical space or a certain area on a larger scale. In fact, in the early days of disaster emergency response, government departments were more concerned about the overall damage rate of the disaster area instead of single-building damage, because this helps the government decide the level of emergency response. We present an innovative algorithm that constructs Neyman stratified random sampling trees for binary classification and extends this approach to multiclass problems. Through extensive experimentation on various datasets and model structures, our findings demonstrate that our method surpasses both passive and conventional active learning techniques in terms of class rate estimation and model enhancement with only 30\%-60\% of the annotation cost of simple sampling. It effectively addresses the 'sampling bias' challenge in traditional active learning strategies and mitigates the 'cold start' dilemma. The efficacy of our approach is further substantiated through application to disaster evaluation tasks using Xview2 Satellite imagery, showcasing its practical utility in real-world contexts.

\end{abstract}



\maketitle

\section{Introduction}

With the rapid development of earth observation technology, we have entered an era of massively available satellite remote-sensing data. However, a large amount of satellite remote sensing data lacks a label~\cite{xu2019building,rudner2019multi3net} or the label cost is too high~\cite{van2018spacenet,bonafilia2020sen1floods11} to hinder the potential of AI technology mining satellite data. Especially in such an emergency response scenario that uses satellite data to evaluate the degree of disaster damage, the lack of labeled data~\cite{Lee2020AssessingPD} has seriously hindered the research and development of the AI model for supporting efficient disaster response. In the past, disaster damage assessment encountered bottlenecks due to excessive focus on the damage of a certain building~\cite{bai2023knowledge,xia2022self} in a specific geographical space or a certain area~\cite{bai2017framework} on a larger scale. This is a challenging problem given the limitations of satellite remote sensing resolution and the development level of AI technology. Indeed, in numerous scenarios, the precision of overall class rate estimations precedes the model's classification accuracy. In fact, in the early days of disaster emergency response, government departments were more concerned about the overall damage rate of the disaster area instead of single-building damage, because this helps the government decide the level of emergency response~\cite{2019Building}.

Although it is difficult to train a classification model with good robustness and high accuracy with limited samples, it is possible to achieve a high confidence class rate estimation (estimating the class rate of population unbiasedly ) result. In settings where labeled data is scarce, the efficient annotation of data poses a substantial challenge. Active learning rises to this challenge by selecting the most informative instances for annotation to reduce total labeling costs. Active learning is an effective method for performing class rate estimation. Active learning methods' mainstream  query strategies are rule-based non-probabilistic sampling methods. However, samples obtained  cannot be directly used for class rate estimation tasks.  Using probability sampling allows for an unbiased estimate of the class rate, with sampling errors decreasing as the magnitude of labeled samples increases. However, traditional probability sampling, equivalent to a "passive learning" process without utilizing model information, cannot effectively guide model learning.

\begin{figure}
    \centering
    \includegraphics[width=0.9\linewidth]{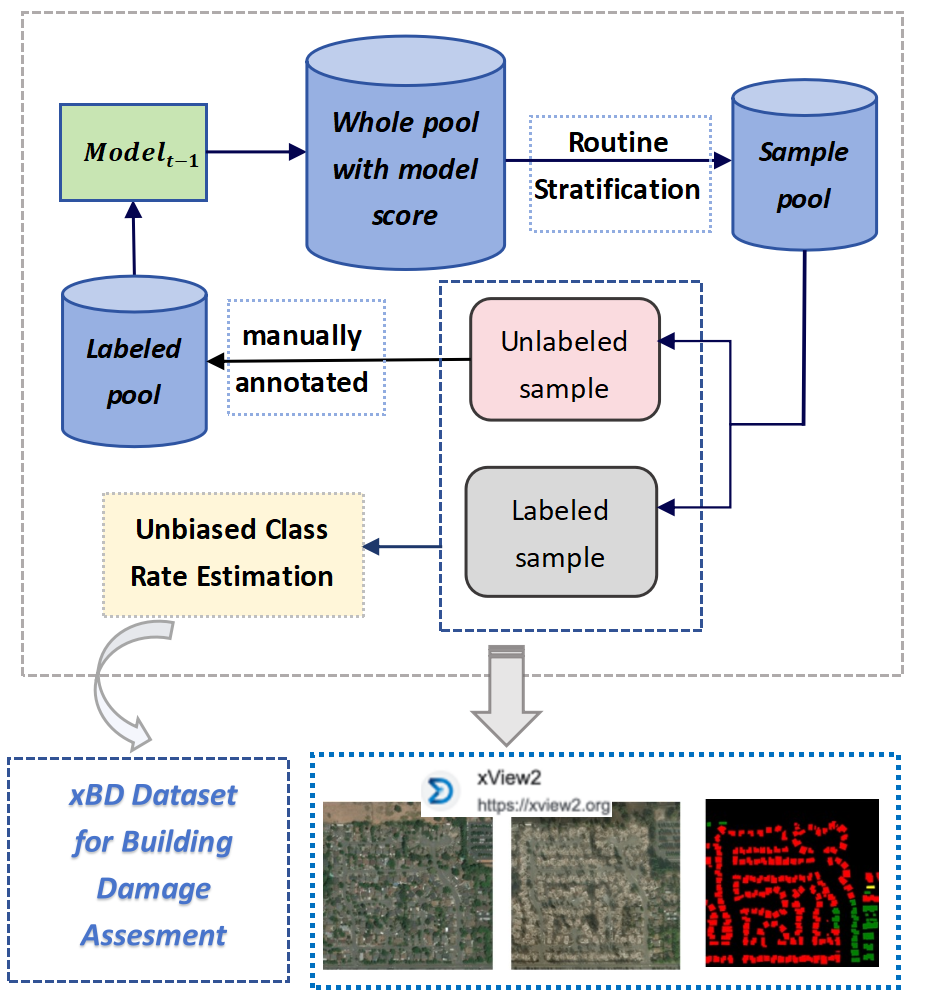}
    \caption{Framework  of the proposed approach}
    \label{lab0}
\end{figure}

\Cref{lab0} shows the research framework of this paper. In this study, We propose an active learning method based on Neyman stratified random sampling as the query strategy which combines the ideas of probability sampling and active learning, leveraging their respective strengths in class rate estimation and model training tasks. It efficiently utilizes labeling costs for annotation, allowing the obtained labeled samples to simultaneously serve model training and class rate estimation tasks. The proposed method is of practical value for many application scenarios that require both model training and high-confidence class rate estimation.

 The paper conducts experimental evaluations on multiple public datasets and model structures to validate the effectiveness of the proposed approach. Our method surpasses both passive and conventional active learning techniques in terms of class rate estimation and model enhancement with only 30\%-60\% of the annotation cost of simple sampling.  Finally, an application example in the context of building damage assessment based on Xview2 satellite remote sensing images in xBD dataset illustrates its practical application in rapid disaster response.


\section{Related Work}
Previous research on building damage assessment mainly focused on classification models of single buildings or regional buildings~\cite{2019xBD,2019Building,xu2019building}, and will not be introduced in detail here.
Concerning the commonly employed uncertainty-based query strategy in active learning, numerous studies have pinpointed sampling bias as a prevalent issue. In response, there is an increasing inclination towards the fusion of diverse query strategies. Yang et al. ~\shortcite{10.1007/s11263-014-0781-x} have synthesized uncertainty with diversity strategies, evaluating data uncertainty across the full spectrum of the active pool. Meanwhile, Chu et al. ~\shortcite{7837913}have been exploring the transferability of active learning, devising a model that integrates various strategies, thereby quantifying the extent of active learning using linear weights.

The field of active learning is increasingly exploring interdisciplinary combinations with other fields. Both active learning and semi-supervised learning share the common objective of improving learning with limited labeled data. Gao et al. ~\shortcite{10.1007/978-3-030-58607-2_30}introduced a comprehensive framework that integrates both approaches. In the context of transfer learning, Xie et al. ~\shortcite{DBLP:journals/corr/abs-2112-01406} introduced an active learning strategy known as "active domain adaptation." This approach utilizes a straightforward energy-based sampling strategy, selecting data based on domain-specific characteristics and model prediction uncertainty.

In active learning, cold-start problem arises if the initial labeled sample set is not large enough. When the cold-start problem occurs, samples based on active learning do not perform better than simple random sampling ~\cite{6889457}. Therefore, the cold-start problem has always been a crucial research direction in the field of active learning. Gao et al. ~\shortcite{10.1007/978-3-030-58607-2_30} adopted a combination of semi-supervised learning and active learning methods to alleviate the issue to some extent.

In the context of class rate estimation, active learning aims to improve the accuracy of probability model predictions for unlabeled samples. Saar-Tsechansky et al.~\shortcite{10.1023/B:MACH.0000011806.12374.c3} proposed BOOTSTRAP-LV, an active learning approach that selects samples with high variance in model output class rate estimates. This method incorporates the accuracy measure of class rate estimates into the active learning query strategy, enhancing both model training accuracy and class rate estimation precision. Melville et al.~\shortcite{10.1007/11564096_28} further improved BOOTSTRAP-LV by using Jensen-Shannon divergence to measure high information content. 

However, relying solely on model-predicted probabilities for overall class rate estimation may not be useful for certain applications, such as disaster assessment based on remote sensing images and internet monitoring scenarios. In these scenarios, exploring how to directly utilize labeled samples acquired through iterative sampling in active learning is a valuable direction. However, due to this perspective not being mainstream in the active learning field, relevant works in this area are limited, and we take this research perspective as our starting point.

\section{Method}
\subsection{Active Learning Ideas in Stratified Sampling}
\label{section3.1}
Stratified Random Sampling with Neyman’s allocation, is a probabilistic sampling method that embodies the uncertainty in active learning. It allocates sample sizes among strata by considering inter-stratum uncertainties. Additionally, it ensures coverage of the entire population distribution, providing every sample in the population a probability of being selected. This aligns with the diversity concept in active learning, helping mitigate the "sampling bias" issue in the uncertain query strategy. 

The following will further compare it with the uncertainty query strategy in active learning, thereby further illustrating the active learning ideas inherent in stratified random sampling with Neyman’s allocation.

For the uncertainty entropy sampling strategy in active learning, it selects a batch of samples with the highest entropy in each iteration. In binary classification, this means collecting more samples with model predictions  around 0.5, near the decision boundary.Assuming a well-calibrated model, the Neyman Stratified Random Sampling strategy allocates fewer samples to extreme model predictions   (close to 0 or 1) and more samples to  around 0.5. This reflects a similar principle between Neyman Stratified Random Sampling and uncertainty entropy sampling, focusing on allocating more samples around the model's decision boundary.

\begin{figure}
    \centering
    \includegraphics[width=1\linewidth]{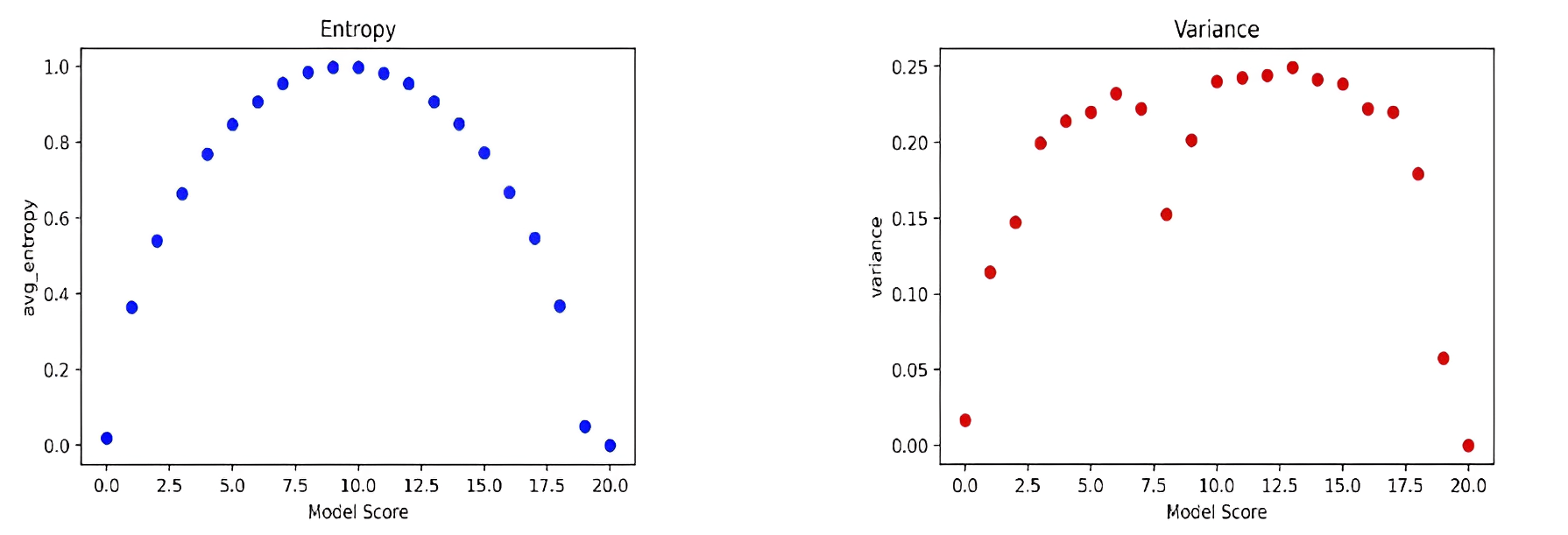}
\captionsetup{justification=raggedright,singlelinecheck=false}
    \caption{Distribution of entropy and variance estimation along the dimension of model score.}
    \label{fig1}
    \Description{}
\end{figure}

Simultaneously, it can be directly observed from the data that the two share similar ideas. \Cref{fig1} shows a similar distribution pattern of sample entropy and estimated variance along the model predicted scores dimension. The ideal variance distribution forms a perfect parabolic shape in a well-calibrated model. Despite practical deviations, the basic trend of variance distribution adheres to a parabolic distribution.

Furthermore, in practice, sample allocation will be performed using the following formula:
\begin{equation}
n_h=\min \left(\max \left(n_{\text {threshold }}, n \frac{W_h S_h}{\sum_{k=1}^L W_k S_k}\right), N_h\right)
\end{equation}
Where the inner max function introduces a minimum threshold for sample allocation to ensure each layer receives a minimum sample allocation. This Neyman stratified random sampling considers both within-layer variance and overall distribution, incorporating the diversity idea from active learning. It prevents samples from being concentrated solely near the decision boundary, addressing the "sampling bias" issue in the uncertainty query strategy. The outer min function ensures that Neyman sample allocation does not exceed the total population within the layer.

\subsection{Main Algorithm}
 The proposed approach incorporates Neyman stratified random sampling as the query strategy into the iterative training process of active learning(details in \Cref{alg:active_learning}).
 
\begin{algorithm}[tb]
    \caption{Main Algorithm of the Proposed Approach}
    \label{alg:active_learning}
    
    \begin{description}
        \item[Input:] Total number of iterations $T$; Initial unlabeled pool size $N$; Initial sampling size $n_{init}$; Sampling size at iteration $t$ $n_{roundt}$; In iteration round t, number of layers $L_t$, Population size in layer h $N_{th}$, Weight of layer h $W_{th}$, 
        Sample mean within layer h  
 $\overline{y_{th}}$, sample variance within layer h ${s_{th}^{2}}$
        
        \item[Initialize (Iteration $t=0$):] Initial simple random sampling $n_{init}$ as the initialization of the labeled training sample pool; Train the initial iteration $Model_0$  based on the initialized labeled training sample pool
    \end{description}
    
    \textbf{Output:} Estimated overall class rate
    
    \begin{algorithmic}[1]
        
            \WHILE{the current unlabeled pool is not empty:}  
                \STATE  Infer the total sample pool (including labeled and unlabeled) using the $Model_{t-1}$ to obtain a sample pool with $Model_{t-1}$ estimated scores.
                \STATE  Perform Neyman stratified random sampling on the sample pool using model estimated scores, obtaining the sample selection results for the current iteration.
                \STATE  Manually label the selected samples 
                \STATE  Based on the labeled results  in the current iteration t, provide an unbiased estimate for the overall class rate.
            $$\widehat{\overline{Y}_{t}}=\sum_{h=1}^{L_{t}}W_{th}\overline{y_{th}}=\frac{1}{N}\sum_{\mathrm{h=1}}^{L_{t}}N_{\mathrm{th}}\overline{y_{th}}$$
                \STATE Update the labeled training sample pool with the newly obtained labeled samples for model training $Model_{t}$.
            \ENDWHILE

    \STATE 
    \textbf{return:} $\widehat{\overline{Y}}=\frac{1}{T}\sum_{t=1}^T\widehat{\overline{Y}_t}$
    \end{algorithmic}
\end{algorithm}

It's important to highlight that, unlike typical active learning , our query strategy samples from the overall pool in each iteration, including both labeled and unlabeled samples. This ensures that Neyman stratified random sampling in each iteration provides an unbiased estimate of the overall class rate. However, a challenge arises as the samples selected by Neyman stratified random sampling may include already labeled ones. To avoid redundant annotations, in step 2 of the \Cref{alg:active_learning}'s iteration, selected samples are deduplicated with the labeled pool before annotation. Afterward, the deduplicated samples are annotated and merged into the labeled training sample pool.

\subsection{Tree Based Routine Stratification}
\label{section3.3}
During the model iteration process in active learning, the distribution of samples on the model estimated scores is continuously updated. Therefore, the stratified sampling scheme should also be iteratively updated. The following provides a routine implementation of Neyman stratified random sampling based on tree models in the iterative process, which allows for the routine determination of "good" stratification schemes for each iteration.
We view Neyman's stratified random sampling as an optimization problem, aiming to minimize the variance of class rate estimation. Since Neyman allocation is not inherently suitable for multi-class problems, we'll first illustrate the optimization problem with a binary classification example, as defined in Equation:

\begin{equation}\begin{cases}x=[x_0=0,x_1,x_2,...,x_{L-1},x_L=1],\\bucket_h=\{i;x_h\leq model(i)\leq x_{h+1}\},\\W_h=\frac{N_h}{N},s_h^2=p_h*(1-p_h),\\ST\quad\forall x_k\in[0,1],\\MIN(\sum_{h=1}^L\frac1{n_h}W_h^2s_h^2)\end{cases}\end{equation}

Where $[x_0=0,x_1,x_2,...,x_{L-1},x_L=1]$ is the stratification threshold points for each layer, $bucket_h$ is the overall set of layer h, $bucket_h$is the layer weights within layer h, $p_h$ is the sample class rate within layer h, and $p_h$ is the sample variance within layer h.
In practical, a binary tree, referred to as a stratified tree, is used for stratified random sampling whose time complexity is $O(logn)$ . The process is specifically explained for binary classification in\Cref{alg:stratified_tree}.
\begin{algorithm}[tb]
    \caption{Stratified Tree Algorithm(iteration round t)}
    \label{alg:stratified_tree}
    \textbf{Initialize}: Total sample pool (including labeled and unlabeled samples in iteration round t) as the root node of the tree\\
    \textbf{Input}: Total sample pool as the training set D; $Model_{t-1}$'s predicted scores as the feature set A\\
    \textbf{Stopping Condition}: Set tree depth as k so the maximum expected number of layers is $2^{k-1}$\\
    \textbf{Output}: Stratified Tree
   
    \begin{algorithmic}[1]
       
        \STATE \textbf{Procedure GenerateStratifiedTree}($D, A$):
       \IF {all instances in $D$ belong to the same class }
        \STATE \textbf{return} tree T
        \ELSIF{$A$ is an empty set}  
        \STATE \textbf{return} tree T
        \ELSE
        \STATE Calculate the target function for splitting
        \STATE Choose the split threshold for features A based on the target function
        \STATE Split the dataset D into two leaf nodes based on the feature A's split threshold
        \STATE Construct  tree $T$ with the current node and  new leaf nodes
        \ENDIF
        \FOR{each child node i in the tree $T$:}
        \STATE  Call \textbf{GenerateStratifiedTree}($D_{\text{i}}, A_{\text{i}}$) to get the subtree $T_i$
        \ENDFOR
        \STATE \textbf{return} tree $T$
    \end{algorithmic}
\end{algorithm}

Note that unlike standard CART trees, the split in Neyman stratified random sampling tree minimizes the sampling error estimate across the entire stratified result, considering both local and global changes in leaf nodes.

\subsection{Solution for Multiclass Classification}
\label{section3.4}
In binary classification, the feature A is a single-dimensional feature, i.e., the $Model_{t-1}$'s estimated probability of positive instances for the samples. The allocation principle in Neyman allocation is optimized for a single class rate estimation target. In the case of multiple classes and objectives, the optimal allocation for one specific target may not necessarily be optimal for other targets. For multi-class problems (like K classification), this paper also provide an corresponding applicable solution:

In multi-class problems, the feature set A is composed of the vector of estimated positive probabilities of samples by the . Firstly, select M classes from the total of K classes, focusing on those more relevant in practical scenarios, as positive classes. The remaining K-M classes are considered negative classes, thus transforming the multi-class problem into a binary classification problem.

However, the obtained optimal Neyman allocation scheme may still not be  applicable to each of the M classes. Therefore, specific guardrail determination rules are established for each of these M classes to ensure that the final sample allocation plan is suitable for each positive class(that is, adding a new decision condition in steps 3 and 4 of the stratified tree algorithm described in \Cref{alg:stratified_tree}). The specific condition is presented as follows:

\begin{equation}\sum_{h=1}^{L}\frac{1}{n_{h}}W_{h}^{2}s_{hm}^{2}\geq\frac{1}{n}s_{m}^{2}\end{equation}
Where the positive class m = 1,…, M. If this decision condition is satisfied, then the node will not split at this threshold point {\it split}, and the algorithm returns a single-node tree T with this node as the root; otherwise, the algorithm proceeds to step 4.

The condition ensures that if the final stratified sampling result has a larger sampling error for positive class {\it m} compared to simple random sampling based on the current threshold point {\it split}, no stratification will be performed for that threshold point. This "guardrail condition" guarantees optimization for each positive class among M, with sampling errors smaller than simple random sampling. Validation of the solution for multiclass classification will be discussed in \Cref{section4.5}.

\section{Experiments}
We conduct extensive experiments across multiple datasets and model structures to compare the effectiveness of our proposed active learning method  based on stratified random sampling with Neyman’s optimum allocation(NSRS) against passive learning method with simple random sampling (SRS) and the traditional active learning method with uncertainty entropy sampling (UES). We use open computer vision datasets, including MNIST, CIFAR-10, and CINIC-10. The model structures include logistic regression, M1 (a custom simple convolutional neural network), Lenet5, Alexnet8, VGG16, and RESNET50. Training parameters involve using the Adam optimizer with an initial learning rate of 0.001 and decay mechanism, along with a batch size of 100.
\subsection{Class Rate Estimation Results}
\label{section4.1}
\Cref{tab1} displays class rate estimates(the overall category rate is 0.1) and variances for three query strategies on the CIFAR-10 dataset, assuming a sampling size of 10,000. Neyman Stratified Random Sampling, based on model predictions, is affected by the quality of model predictions. Results are presented for logistic regression, M1, and Alexnet8 models, with variations based on model complexity and training iterations(represented as cumulative training sets).

\begin{table}
  \centering  
    \begin{tabular}{llrrrrrrrrrrrr}
    \toprule
     &\multicolumn{6}{r}{Cumulative Training Volume}\\
     &Query & Statistics& 20\% & 40\% & 60\% & 80\% \\
    
    \midrule
    LR & SRS & Estimate & 0.1 & 0.1 & 0.1 & 0.1 \\
          &       & V($10^{-6}$) & 9.00 & 9.00 & 9.00 & 9.00 \\
    & NSRS & Estimate & 0.097 & 0.098 & 0.101 & 0.099 \\
          &       & V($10^{-6}$) & 6.17 & 6.02 & 5.79 & 5.51 \\
          & UES & Estimate & 0.268 & 0.201 & 0.152 & 0.121 \\
        
    M1 & SRS & Estimate & 0.1 & 0.1 & 0.1 & 0.1 \\
          &       & V($10^{-6}$) & 9.00 & 9.00 & 9.00 & 9.00 \\
    & NSRS & Estimate & 0.1 & 0.1 & 0.098 & 0.1 \\
          &       & V($10^{-6}$) & 5.28 & 2.76 & 1.92 & 1.70 \\
          & UES & Estimate & 0.372 & 0.192 & 0.124 & 0.09 \\
   
    A8 & SRS & Estimate & 0.1 & 0.1 & 0.1 & 0.1 \\
          &       & V($10^{-6}$) & 9.00 & 9.00 & 9.00 & 9.00 \\
    & NSRS & Estimate & 0.098 & 0.104 & 0.1 & 0.1 \\
          &       & V($10^{-6}$) & 2.65 & 3.64 & 1.95 & 1.40 \\
          & UES & Estimate & 0.375 & 0.246 & 0.166 & 0.125 \\
    
    \bottomrule
    
    \end{tabular}
    \captionsetup{justification=raggedright,singlelinecheck=false} 
    \caption{Comparison of Class Rate Estimation Results. Where LR represents logistic regression model, M1 represents M1 model, A8 represents  Alexnet8 model, V represents variance of estimate}
   \label{tab1}
\end{table}

\begin{figure}
    \centering
    \includegraphics[width=1\linewidth]{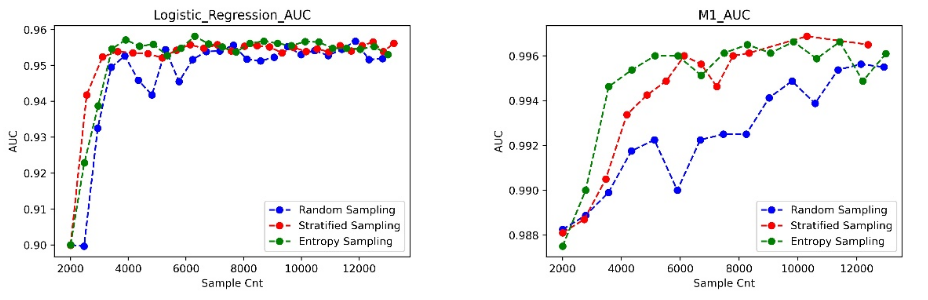}
    \captionsetup{justification=raggedright,singlelinecheck=false}
    \caption{Comparison of model performance with three query strategies on the MNIST dataset}
    \label{fig2}
    \Description{}
\end{figure}

In \Cref{tab1}, we can find that both Simple Random Sampling and Neyman Stratified Random Sampling yield unbiased estimates around the true value of 0.1, whereas uncertainty entropy sampling does not. Neyman Stratified Random Sampling shows 30\%-60\% lower variance compared to simple random sampling. The variance decreases with increased model complexity and cumulative training volume, indicating improved performance. This indicates that the quality of model predictions influences the effectiveness of stratification. Better model prediction results in improved performance of  the proposed approach based on the estimated model scores.

In summary,  the proposed approach allows unbiased overall class rate estimation through iterative sampling, a capability lacking in traditional uncertainty entropy sampling. Additionally, it achieves lower variance, requiring only 30\%-60\% of the labeling cost compared to simple random sampling for equivalent sampling error. This holds practical value for applications relying on sampling annotation for class rate estimation.

\subsection{Model Training Results}

\begin{figure}
    \centering
    \includegraphics[width=1\linewidth]{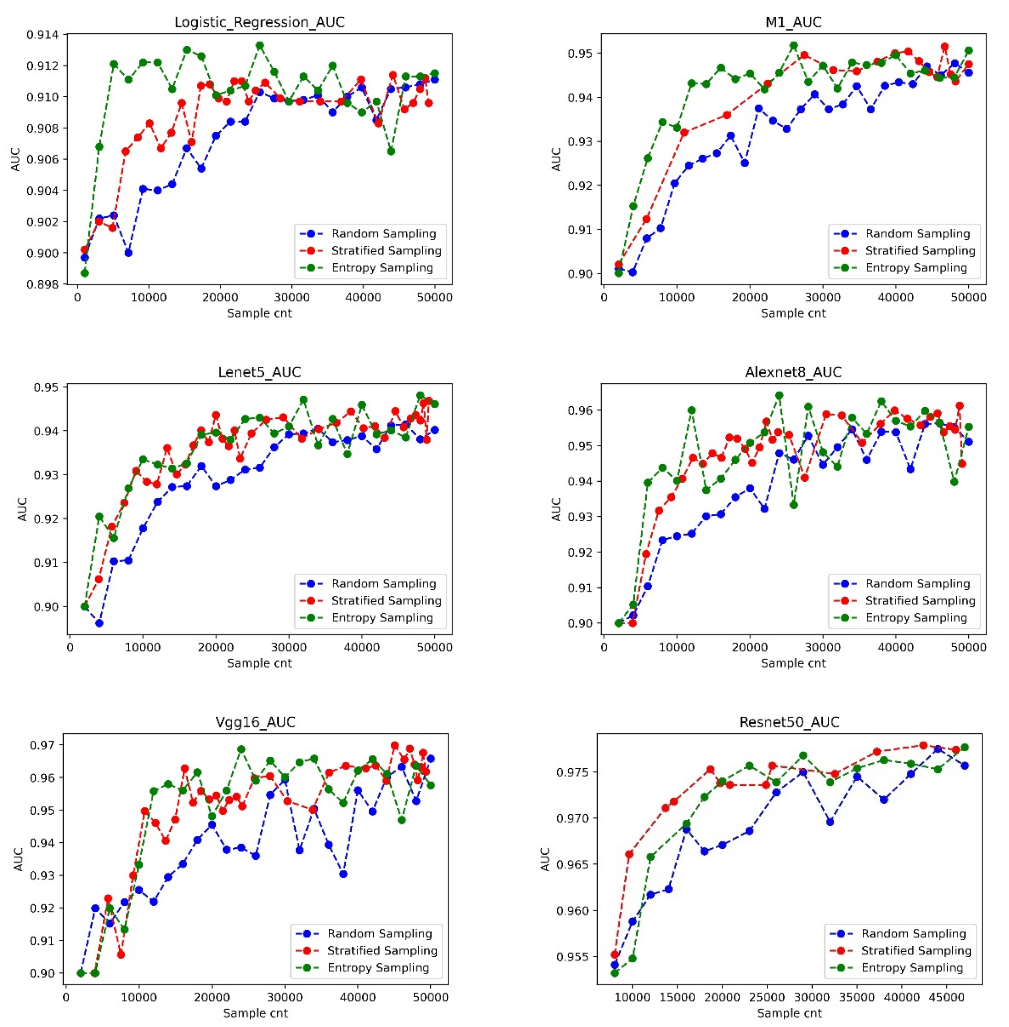}
    \captionsetup{justification=raggedright,singlelinecheck=false}
    \caption{Comparison of model performance with three query strategies on the CIFAR-10 dataset}
    \label{fig3}
    \Description{}
\end{figure}

\begin{figure}
    \centering
    \includegraphics[width=1\linewidth]{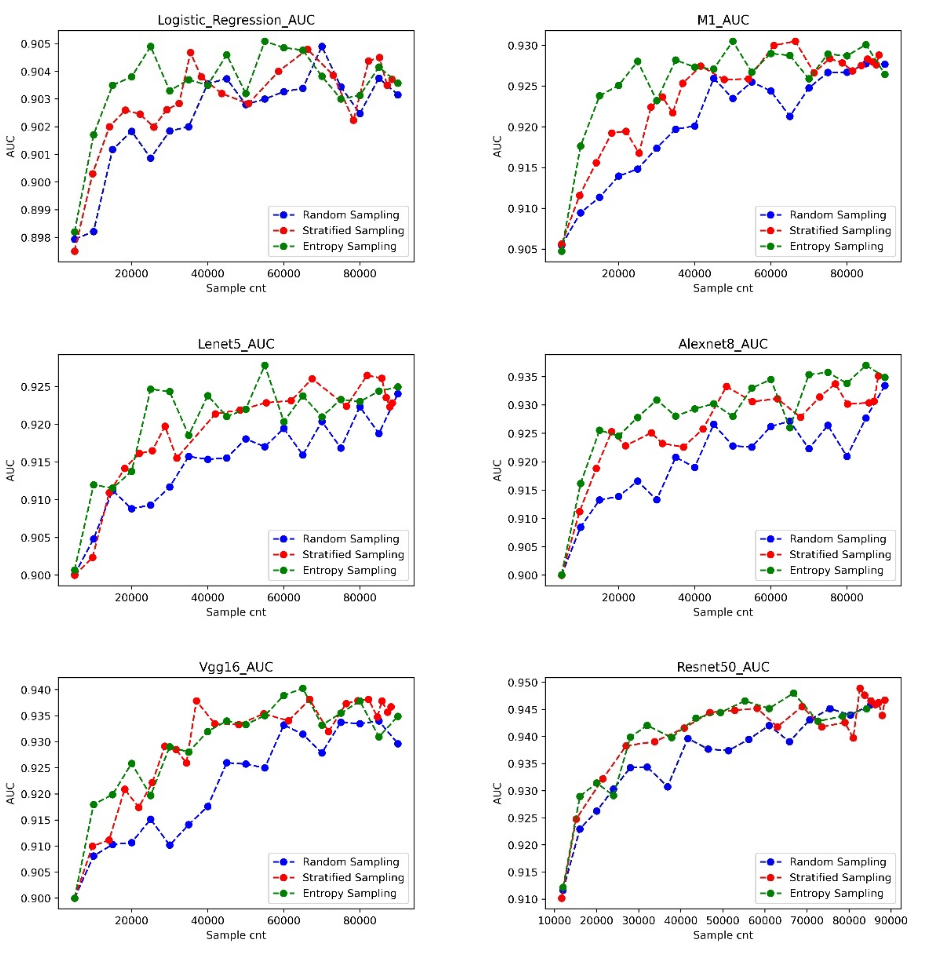}
    \captionsetup{justification=raggedright,singlelinecheck=false}
    \caption{Comparison of model performance with three query strategies on the CINIC-10 dataset}
    \label{fig4}
    \Description{}
\end{figure}

\Cref{fig2}, \ref{fig3}, and \ref{fig4} depict AUC curve results for model iteration and training across three datasets (MNIST, CIFAR-10, CINIC-10) using three query strategies (Neyman stratified random sampling, simple random sampling, uncertainty entropy sampling) on six model structures (logistic regression, M1, Lenet5, Alnex8, Vgg16, Resnet50). The x-axis represents training sample size, and the y-axis represents model AUC results. Notably, due to MNIST's simplicity, experiments focused on logistic regression and M1 models with a smaller sample size.

Neyman stratified random sampling outperforms simple random sampling across datasets and models, reflecting its active learning ideas. This emphasizes its ability to choose good samples based on model information during iteration. Neyman stratified random sampling and uncertainty entropy sampling show comparable results on high-complexity models. On low-complexity models, Neyman stratified random sampling slightly falls behind uncertainty entropy sampling, attributed to its dual focus on class rate estimation and model training compared to the latter's emphasis on model training. The performance gap diminishes with increasing model complexity, with Neyman stratified random sampling slightly outperforming on Resnet50. This is because Neyman stratified random sampling alleviates the "sampling bias" issue caused by uncertainty entropy sampling, leveraging overall distribution information.

\subsection{Sampling Distribution}
\begin{figure}
    \centering
    \includegraphics[width=1\linewidth]{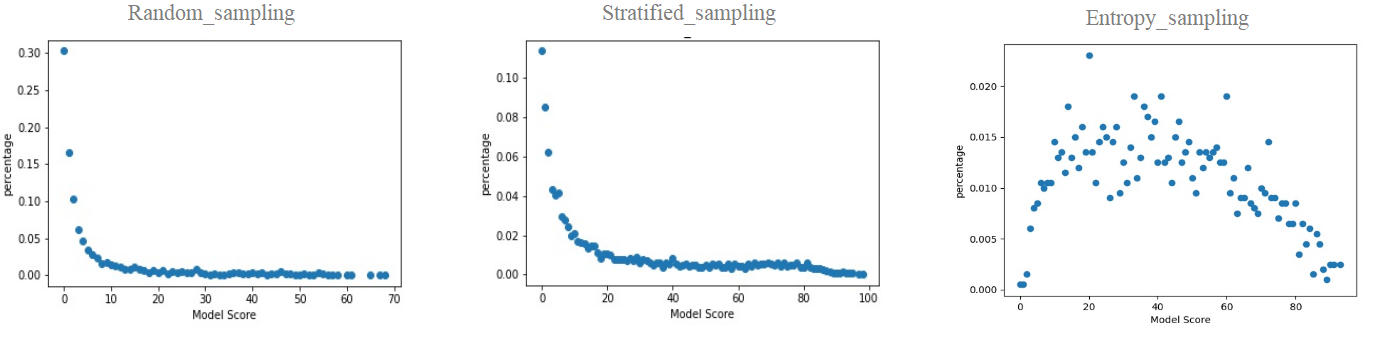}
    \captionsetup{justification=raggedright,singlelinecheck=false}
    \caption{Sample distribution of three query strategies (simple random sampling, Neyman stratified random sampling, uncertainty entropy sampling)}
    \label{fig5}
    \Description{}
\end{figure}

For a more visual comparison of the three sampling strategies, \Cref{fig5} illustrates the distribution of samples based on the model's estimated scores in the current round.

From the figure, Neyman stratified random sampling shows a distribution similar to simple random sampling, closely aligning with the overall sample distribution. Notably, Neyman stratified random sampling samples more around the model's estimated score of 0.5 and less in areas close to 0 or 1. This aligns with the active learning ideas discussed in \Cref{section3.1}.

In contrast, uncertainty entropy sampling concentrates samples around the model's estimated score of 0.5, indicating a reliance on the current model's estimation . Neyman stratified random sampling and uncertainty entropy sampling share similar principles and Neyman stratified random sampling can mitigate the "sampling bias" issue of  uncertainty entropy sampling to some extent.

\subsection{Cold-start problem}
\begin{figure}
    \centering
    \includegraphics[width=1\linewidth]{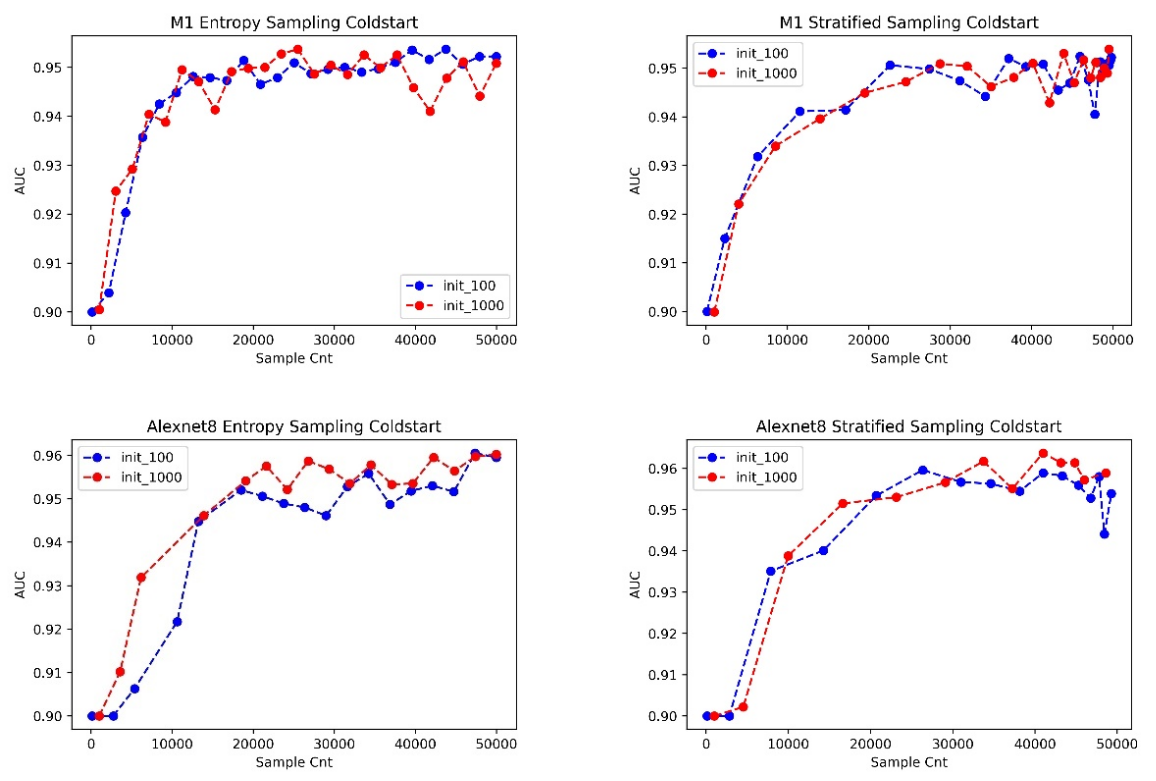}
\captionsetup{justification=raggedright,singlelinecheck=false}
    \caption{Comparison of Sensitivity to Initial Sets Between Neyman Stratified Sampling and Uncertainty Entropy Sampling}
    \label{fig6}
    \Description{}
\end{figure}

To some extent, Neyman stratified random sampling can mitigate the cold-start problem observed in uncertainty entropy sampling. In contrast to uncertainty entropy sampling, the query strategy based on Neyman stratified random sampling exhibits better robustness to the initial set size, as illustrated in \Cref{fig6} on the CIFAR-10 dataset using two different model structures (M1, Alexnet8) with different initial set sizes (100 and 1000).

Uncertainty entropy sampling exhibits a cold-start problem, particularly in more complex networks. When the initial set size is 100, there is a notable drop in the improvement of model AUC after first iteration, a trend that becomes more pronounced in deeper network structures. On the Alexnet8 network, the model's iterative training performance with an initial set of 100 is significantly inferior to that with an set of 1000. As the network complexity increases, Neyman stratified random sampling also encounters a cold-start problem. In the Alexnet8 network structure, Neyman stratified random sampling demonstrates a similar pattern where the improvement in model AUC in the first iteration is noticeably higher than in subsequent iterations. However, overall, Neyman stratified random sampling query strategy proves more effective than uncertainty entropy sampling in addressing the cold-start problem.

\subsection{Multiclass Classification Problem}
\label{section4.5}
This section validates the effectiveness of Neyman stratified random sampling for multi-class classification, comparing three query strategies on CIFAR-10. The dataset focuses on "motorcycle" and "airplane" as positive instances (M=2), merging the remaining eight classes into a single category for a three-class dataset (K=3). Training is performed using the Alexnet8 model.

\begin{table}
  \centering  
    \begin{tabular}{llrrrrrrrrrrrr}
    \toprule
     &\multicolumn{6}{r}{Cumulative Training Volume}\\
     &Query & Statistics& 20\% & 40\% & 60\% & 80\% \\
    
    \midrule
    M & SRS & Estimate & 0.1 & 0.1 & 0.1 & 0.1 \\
          &       & V($10^{-6}$) & 9.00 & 9.00 & 9.00 & 9.00 \\
    & NSRS & Estimate & 0.102 & 0.102 & 0.101 & 0.101 \\
          &       & V($10^{-6}$) & 5.30 & 4.19 & 4.24 & 4.18 \\
          & UES & Estimate & 0.20 & 0.18 & 0.16 & 0.12 \\
     
    A & SRS & Estimate & 0.1 & 0.1 & 0.1 & 0.1 \\
          &       & V($10^{-6}$) & 9.00 & 9.00 & 9.00 & 9.00 \\
    & NSRS & Estimate & 0.102 & 0.101 & 0.10 & 0.10 \\
          &       & V($10^{-6}$) & 7.04 & 5.61 & 5.58 & 5.05 \\
          & UES & Estimate & 0.16 & 0.19 & 0.16 & 0.12 \\

    \bottomrule
    
    \end{tabular}
    \captionsetup{justification=raggedright,singlelinecheck=false} 
    \caption{Comparison of Class Rate Estimation in Multiclass Classification Experiments. Where M represents motorcycle, A represents airplane, V represents variance of estimate.}
   \label{tab2}
\end{table}

\Cref{tab2} summarizes class rate estimation in multi-class classification problems, echoing \Cref{section4.1}. Simple random sampling and Neyman stratified sampling yield unbiased estimates around 0.1, while uncertainty entropy sampling lacks unbiasedness. Neyman stratified sampling shows lower variances on positive classes. Its optimization effect improves with cumulative training volume increasing. In multi-class problems, Neyman stratified sampling's optimization effect is slightly lower than in binary classification. These results affirm the effectiveness of Neyman stratified random sampling for multi-class classification, aligning with \Cref{section3.4}'s theory—it optimizes each positive class target individually but doesn't guarantee optimal results for every target.

\begin{figure}
    \centering
    \includegraphics[width=0.8\linewidth]{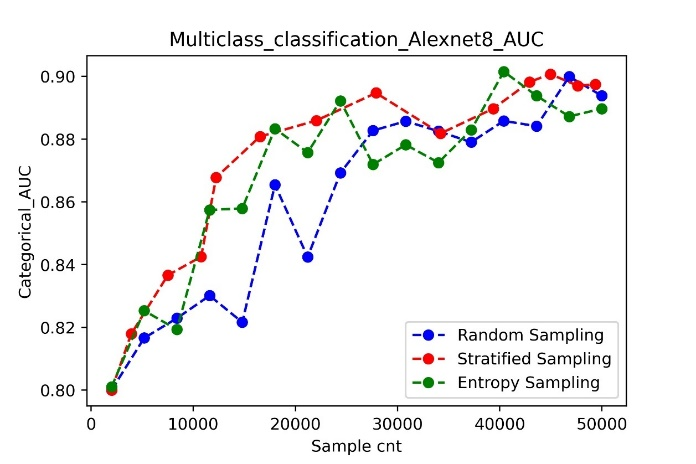}
    \captionsetup{justification=raggedright,singlelinecheck=false}
    \caption{Comparison of Model Performance of Three Query Strategies in multi-class Classification Problem}
    \label{fig7}
    \Description{}
\end{figure}

\subsection{Application }
In this section, we illustrate the practical application of  the proposed approach in disaster assessment using real xBD dataset. The dataset consists of 22,068 color images labeled with 19 events, suitable for disaster localization and assessment~\cite{2019xBD}. We use the Train dataset, dividing it into a 4:1 ratio for training and testing. We extract building polygons from each original image, resizing them to 32 pixels * 32 pixels for the training set. Focusing on Hurricane Harvey and Hurricane Matthew data, we perform a 4-class disaster assessment using Alexnet8 as the base model (K=4). The positive classes (M=3) include Minor damage, Major damage, and Destroyed.

\Cref{fig7} shows a notable improvement with Neyman stratified random sampling compared to the baseline of simple random sampling. It also slightly outperforms uncertainty entropy sampling.

\begin{figure}
    \centering
    \includegraphics[width=1\linewidth]{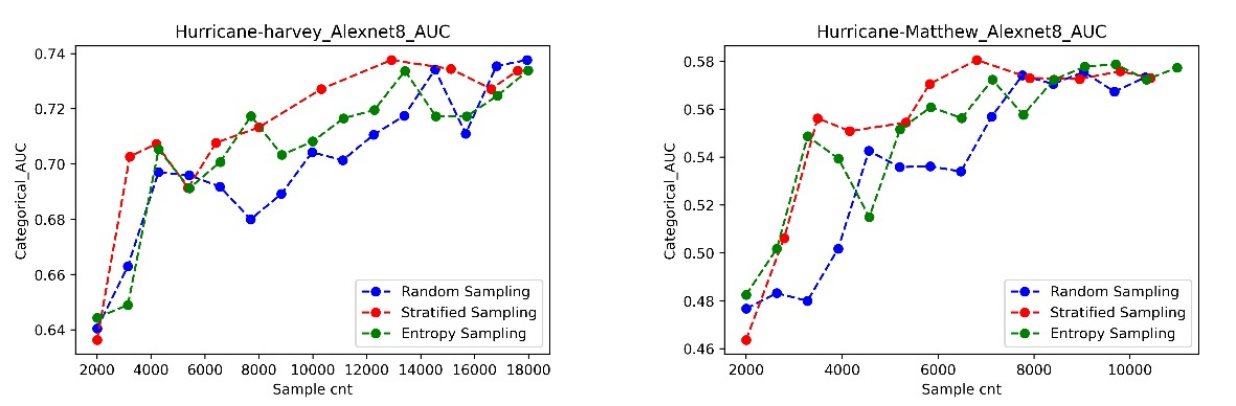}
    \captionsetup{justification=raggedright,singlelinecheck=false}
    \caption{Classification Effect of Three Query Strategies on the xBD Dataset}
    \Description{}
    \label{fig8}
\end{figure}

The result shows the proposed approach's sample selection during model training provides unbiased estimates of true damage rates, with estimated variance 80\%-90\% lower than simple random sampling. Although the optimization degree is slightly lower than on open datasets, attributed to the greater difficulty and lower accuracy of real datasets. In \Cref{fig8},  the proposed approach significantly outperforms passive learning with simple random sampling and slightly surpasses the traditional active learning method with uncertainty entropy strategy in two disaster events. 

Overall,  the proposed approach has practical value. It enables rapid acquisition of unbiased estimates with high confidence, aiding rescue personnel in promptly making initial plans. Moreover, collected samples swiftly enhance model performance for more extensive disaster assessments, supporting further rescue planning.

\section{Conclusion}

This paper introduces an active learning method that combines active learning and probability sampling,  using stratified random sampling with Neyman allocation, minimizing the annotation costs while addressing class rate estimation and model training tasks .

In class rate estimation, the proposed approach provides unbiased estimation of the overall class rate with  lower variance compared to simple random sampling. This allows for equivalent results with only 30\%-60\% of the annotation cost, whcih has great application value in  scenarios that rely on sample labeling. The proposed method shows notable improvement in model training compared to passive learning with simple random sampling. However, it slightly falls behind the traditional active learning strategy (uncertainty entropy sampling), as the proposed method  addresses both class rate estimation and model training. As model complexity increases, the performance gap narrows .
The paper shows, both theoretically and through experiments, that the proposed method, with its inclusion of overall distribution information, can alleviate "sampling bias." In practical scenarios, the method performs well in the cold-start problem. Its real-world applicability is evident in maintaining great performance on the xBD dataset.

The study identified limitations and future prospects of the proposed method. The implementation of the Neyman stratified random sampling query strategy relies on model prediction scores as the stratification dimension in each iteration. Therefore,  the proposed approach is only suitable for explicit classification models that output model probabilities. The study acknowledges that  the proposed approach and traditional uncertainty-based active learning methods each have their advantages, suggesting future research directions that explore the fusion of these two approaches.

\clearpage

\bibliographystyle{ACM-Reference-Format}
\bibliography{sample-base}


\begin{thebibliography}{46}


\ifx \showCODEN    \undefined \def \showCODEN     #1{\unskip}     \fi
\ifx \showDOI      \undefined \def \showDOI       #1{#1}\fi
\ifx \showISBNx    \undefined \def \showISBNx     #1{\unskip}     \fi
\ifx \showISBNxiii \undefined \def \showISBNxiii  #1{\unskip}     \fi
\ifx \showISSN     \undefined \def \showISSN      #1{\unskip}     \fi
\ifx \showLCCN     \undefined \def \showLCCN      #1{\unskip}     \fi
\ifx \shownote     \undefined \def \shownote      #1{#1}          \fi
\ifx \showarticletitle \undefined \def \showarticletitle #1{#1}   \fi
\ifx \showURL      \undefined \def \showURL       {\relax}        \fi
\providecommand\bibfield[2]{#2}
\providecommand\bibinfo[2]{#2}
\providecommand\natexlab[1]{#1}
\providecommand\showeprint[2][]{arXiv:#2}

\bibitem[Angluin(1988)]%
        {1988Queries}
\bibfield{author}{\bibinfo{person}{Dana Angluin}.} \bibinfo{year}{1988}\natexlab{}.
\newblock \showarticletitle{Queries and Concept Learning}.
\newblock \bibinfo{journal}{\emph{Machine Learning}} \bibinfo{volume}{2}, \bibinfo{number}{4} (\bibinfo{year}{1988}), \bibinfo{pages}{319--342}.
\newblock


\bibitem[Bai et~al\mbox{.}(2017)]%
        {bai2017framework}
\bibfield{author}{\bibinfo{person}{Yanbing Bai}, \bibinfo{person}{Chang Gao}, \bibinfo{person}{Sameer Singh}, \bibinfo{person}{Magaly Koch}, \bibinfo{person}{Bruno Adriano}, \bibinfo{person}{Erick Mas}, {and} \bibinfo{person}{Shunichi Koshimura}.} \bibinfo{year}{2017}\natexlab{}.
\newblock \showarticletitle{A framework of rapid regional tsunami damage recognition from post-event TerraSAR-X imagery using deep neural networks}.
\newblock \bibinfo{journal}{\emph{IEEE Geoscience and Remote Sensing Letters}} \bibinfo{volume}{15}, \bibinfo{number}{1} (\bibinfo{year}{2017}), \bibinfo{pages}{43--47}.
\newblock


\bibitem[Bai et~al\mbox{.}(2023)]%
        {bai2023knowledge}
\bibfield{author}{\bibinfo{person}{Yanbing Bai}, \bibinfo{person}{Jinhua Su}, \bibinfo{person}{Yulong Zou}, {and} \bibinfo{person}{Bruno Adriano}.} \bibinfo{year}{2023}\natexlab{}.
\newblock \showarticletitle{Knowledge distillation based lightweight building damage assessment using satellite imagery of natural disasters}.
\newblock \bibinfo{journal}{\emph{GeoInformatica}} \bibinfo{volume}{27}, \bibinfo{number}{2} (\bibinfo{year}{2023}), \bibinfo{pages}{237--261}.
\newblock


\bibitem[Balcan et~al\mbox{.}(2007)]%
        {2007Margin}
\bibfield{author}{\bibinfo{person}{Maria~Florina Balcan}, \bibinfo{person}{Andrei~Z. Broder}, {and} \bibinfo{person}{Tong Zhang}.} \bibinfo{year}{2007}\natexlab{}.
\newblock \showarticletitle{Margin Based Active Learning}.
\newblock \bibinfo{journal}{\emph{lecture notes in computer science}} (\bibinfo{year}{2007}).
\newblock


\bibitem[Bonafilia et~al\mbox{.}(2020)]%
        {bonafilia2020sen1floods11}
\bibfield{author}{\bibinfo{person}{Derrick Bonafilia}, \bibinfo{person}{Beth Tellman}, \bibinfo{person}{Tyler Anderson}, {and} \bibinfo{person}{Erica Issenberg}.} \bibinfo{year}{2020}\natexlab{}.
\newblock \showarticletitle{Sen1Floods11: A georeferenced dataset to train and test deep learning flood algorithms for sentinel-1}. In \bibinfo{booktitle}{\emph{Proceedings of the IEEE/CVF Conference on Computer Vision and Pattern Recognition Workshops}}. \bibinfo{pages}{210--211}.
\newblock


\bibitem[Chu and Lin(2016)]%
        {7837913}
\bibfield{author}{\bibinfo{person}{Hong-Min Chu} {and} \bibinfo{person}{Hsuan-Tien Lin}.} \bibinfo{year}{2016}\natexlab{}.
\newblock \showarticletitle{Can Active Learning Experience Be Transferred?}. In \bibinfo{booktitle}{\emph{2016 IEEE 16th International Conference on Data Mining (ICDM)}}. \bibinfo{pages}{841--846}.
\newblock
\urldef\tempurl%
\url{https://doi.org/10.1109/ICDM.2016.0100}
\showDOI{\tempurl}


\bibitem[Cohn et~al\mbox{.}(1994)]%
        {1994Improving}
\bibfield{author}{\bibinfo{person}{David Cohn}, \bibinfo{person}{Les Atlas}, {and} \bibinfo{person}{Richard Ladner}.} \bibinfo{year}{1994}\natexlab{}.
\newblock \showarticletitle{Improving generalization with active learning}.
\newblock \bibinfo{journal}{\emph{Machine Learning}} \bibinfo{number}{15-2} (\bibinfo{year}{1994}).
\newblock


\bibitem[Cohn et~al\mbox{.}(1996)]%
        {10.5555/1622737.1622744}
\bibfield{author}{\bibinfo{person}{David~A. Cohn}, \bibinfo{person}{Zoubin Ghahramani}, {and} \bibinfo{person}{Michael~I. Jordan}.} \bibinfo{year}{1996}\natexlab{}.
\newblock \showarticletitle{Active learning with statistical models}.
\newblock \bibinfo{journal}{\emph{J. Artif. Int. Res.}} \bibinfo{volume}{4}, \bibinfo{number}{1} (\bibinfo{date}{mar} \bibinfo{year}{1996}), \bibinfo{pages}{129–145}.
\newblock
\showISSN{1076-9757}


\bibitem[Dasgupta and Hsu(2008)]%
        {2008Hierarchical}
\bibfield{author}{\bibinfo{person}{Sanjoy Dasgupta} {and} \bibinfo{person}{Daniel Hsu}.} \bibinfo{year}{2008}\natexlab{}.
\newblock \showarticletitle{Hierarchical sampling for active learning}. In \bibinfo{booktitle}{\emph{Machine Learning, Proceedings of the Twenty-Fifth International Conference (ICML 2008), Helsinki, Finland, June 5-9, 2008}}.
\newblock


\bibitem[Gao et~al\mbox{.}(2020)]%
        {10.1007/978-3-030-58607-2_30}
\bibfield{author}{\bibinfo{person}{Mingfei Gao}, \bibinfo{person}{Zizhao Zhang}, \bibinfo{person}{Guo Yu}, \bibinfo{person}{Sercan~\"{O}. Ar\i{}k}, \bibinfo{person}{Larry~S. Davis}, {and} \bibinfo{person}{Tomas Pfister}.} \bibinfo{year}{2020}\natexlab{}.
\newblock \showarticletitle{Consistency-Based Semi-Supervised Active Learning: Towards Minimizing Labeling Cost}. In \bibinfo{booktitle}{\emph{Computer Vision – ECCV 2020: 16th European Conference, Glasgow, UK, August 23–28, 2020, Proceedings, Part X}} (Glasgow, United Kingdom). \bibinfo{publisher}{Springer-Verlag}, \bibinfo{address}{Berlin, Heidelberg}, \bibinfo{pages}{510–526}.
\newblock
\showISBNx{978-3-030-58606-5}
\urldef\tempurl%
\url{https://doi.org/10.1007/978-3-030-58607-2_30}
\showDOI{\tempurl}


\bibitem[Geng et~al\mbox{.}(2008)]%
        {2008Unbiased}
\bibfield{author}{\bibinfo{person}{Bo Geng}, \bibinfo{person}{Linjun Yang}, \bibinfo{person}{Zheng~Jun Zha}, \bibinfo{person}{Chao Xu}, {and} \bibinfo{person}{Xian~Sheng Hua}.} \bibinfo{year}{2008}\natexlab{}.
\newblock \showarticletitle{Unbiased active learning for image retrieval}.
\newblock \bibinfo{journal}{\emph{IEEE Computer Society}} (\bibinfo{year}{2008}).
\newblock


\bibitem[Guo et~al\mbox{.}(2017)]%
        {10.5555/3305381.3305518}
\bibfield{author}{\bibinfo{person}{Chuan Guo}, \bibinfo{person}{Geoff Pleiss}, \bibinfo{person}{Yu Sun}, {and} \bibinfo{person}{Kilian~Q. Weinberger}.} \bibinfo{year}{2017}\natexlab{}.
\newblock \showarticletitle{On Calibration of Modern Neural Networks}. In \bibinfo{booktitle}{\emph{Proceedings of the 34th International Conference on Machine Learning - Volume 70}} (Sydney, NSW, Australia) \emph{(\bibinfo{series}{ICML'17})}. \bibinfo{publisher}{JMLR.org}, \bibinfo{pages}{1321–1330}.
\newblock


\bibitem[Gupta et~al\mbox{.}(2019)]%
        {2019xBD}
\bibfield{author}{\bibinfo{person}{Ritwik Gupta}, \bibinfo{person}{Richard Hosfelt}, \bibinfo{person}{Sandra Sajeev}, \bibinfo{person}{Nirav Patel}, \bibinfo{person}{Bryce Goodman}, \bibinfo{person}{Jigar Doshi}, \bibinfo{person}{Eric Heim}, \bibinfo{person}{Howie Choset}, {and} \bibinfo{person}{Matthew Gaston}.} \bibinfo{year}{2019}\natexlab{}.
\newblock \showarticletitle{xBD: A Dataset for Assessing Building Damage from Satellite Imagery}.
\newblock  (\bibinfo{year}{2019}).
\newblock


\bibitem[Hauptmann et~al\mbox{.}(2006)]%
        {2006Extreme}
\bibfield{author}{\bibinfo{person}{Alexander~G. Hauptmann}, \bibinfo{person}{Wei~Hao Lin}, \bibinfo{person}{Rong Yan}, \bibinfo{person}{Jun Yang}, {and} \bibinfo{person}{Ming~Yu Chen}.} \bibinfo{year}{2006}\natexlab{}.
\newblock \showarticletitle{Extreme video retrieval: joint maximization of human and computer performance}. In \bibinfo{booktitle}{\emph{Acm International Conference on Multimedia}}.
\newblock


\bibitem[He et~al\mbox{.}(2016)]%
        {2016Deep}
\bibfield{author}{\bibinfo{person}{Kaiming He}, \bibinfo{person}{Xiangyu Zhang}, \bibinfo{person}{Shaoqing Ren}, {and} \bibinfo{person}{Jian Sun}.} \bibinfo{year}{2016}\natexlab{}.
\newblock \showarticletitle{Deep Residual Learning for Image Recognition}. In \bibinfo{booktitle}{\emph{IEEE Conference on Computer Vision and Pattern Recognition}}.
\newblock


\bibitem[Hoi et~al\mbox{.}(2006)]%
        {2006Batch}
\bibfield{author}{\bibinfo{person}{Steven C.~H. Hoi}, \bibinfo{person}{Rong Jin}, \bibinfo{person}{Jianke Zhu}, {and} \bibinfo{person}{Michael~R. Lyu}.} \bibinfo{year}{2006}\natexlab{}.
\newblock \showarticletitle{Batch Mode Active Learning and Its Application to Medical Image Classiflcation}. In \bibinfo{booktitle}{\emph{Machine Learning, Twenty-third International Conference, Pittsburgh, Pennsylvania, Usa, June}}.
\newblock


\bibitem[Joshi et~al\mbox{.}(2009)]%
        {5206627}
\bibfield{author}{\bibinfo{person}{Ajay~J. Joshi}, \bibinfo{person}{Fatih Porikli}, {and} \bibinfo{person}{Nikolaos Papanikolopoulos}.} \bibinfo{year}{2009}\natexlab{}.
\newblock \showarticletitle{Multi-class active learning for image classification}. In \bibinfo{booktitle}{\emph{2009 IEEE Conference on Computer Vision and Pattern Recognition}}. \bibinfo{pages}{2372--2379}.
\newblock
\urldef\tempurl%
\url{https://doi.org/10.1109/CVPR.2009.5206627}
\showDOI{\tempurl}


\bibitem[Klidbary et~al\mbox{.}(2017)]%
        {2017Outlier}
\bibfield{author}{\bibinfo{person}{Sajad~Haghzad Klidbary}, \bibinfo{person}{Saeed~Bagheri Shouraki}, \bibinfo{person}{Aboozar Ghaffari}, {and} \bibinfo{person}{Soroush~Sheikhpour Kourabbaslou}.} \bibinfo{year}{2017}\natexlab{}.
\newblock \showarticletitle{Outlier robust fuzzy active learning method (ALM)}. In \bibinfo{booktitle}{\emph{2017 7th International Conference on Computer and Knowledge Engineering (ICCKE)}}.
\newblock


\bibitem[Konyushkova et~al\mbox{.}(2017)]%
        {10.5555/3294996.3295177}
\bibfield{author}{\bibinfo{person}{Ksenia Konyushkova}, \bibinfo{person}{Sznitman Raphael}, {and} \bibinfo{person}{Pascal Fua}.} \bibinfo{year}{2017}\natexlab{}.
\newblock \showarticletitle{Learning Active Learning from Data}. In \bibinfo{booktitle}{\emph{Proceedings of the 31st International Conference on Neural Information Processing Systems}} (Long Beach, California, USA) \emph{(\bibinfo{series}{NIPS'17})}. \bibinfo{publisher}{Curran Associates Inc.}, \bibinfo{address}{Red Hook, NY, USA}, \bibinfo{pages}{4228–4238}.
\newblock
\showISBNx{9781510860964}


\bibitem[Krizhevsky et~al\mbox{.}(2017)]%
        {10.1145/3065386}
\bibfield{author}{\bibinfo{person}{Alex Krizhevsky}, \bibinfo{person}{Ilya Sutskever}, {and} \bibinfo{person}{Geoffrey~E. Hinton}.} \bibinfo{year}{2017}\natexlab{}.
\newblock \showarticletitle{ImageNet Classification with Deep Convolutional Neural Networks}.
\newblock \bibinfo{journal}{\emph{Commun. ACM}} \bibinfo{volume}{60}, \bibinfo{number}{6} (\bibinfo{date}{may} \bibinfo{year}{2017}), \bibinfo{pages}{84–90}.
\newblock
\showISSN{0001-0782}
\urldef\tempurl%
\url{https://doi.org/10.1145/3065386}
\showDOI{\tempurl}


\bibitem[Lecun et~al\mbox{.}(1998)]%
        {726791}
\bibfield{author}{\bibinfo{person}{Y. Lecun}, \bibinfo{person}{L. Bottou}, \bibinfo{person}{Y. Bengio}, {and} \bibinfo{person}{P. Haffner}.} \bibinfo{year}{1998}\natexlab{}.
\newblock \showarticletitle{Gradient-based learning applied to document recognition}.
\newblock \bibinfo{journal}{\emph{Proc. IEEE}} \bibinfo{volume}{86}, \bibinfo{number}{11} (\bibinfo{year}{1998}), \bibinfo{pages}{2278--2324}.
\newblock
\urldef\tempurl%
\url{https://doi.org/10.1109/5.726791}
\showDOI{\tempurl}


\bibitem[Lee et~al\mbox{.}(2020)]%
        {Lee2020AssessingPD}
\bibfield{author}{\bibinfo{person}{Jihyeon Lee}, \bibinfo{person}{Joseph~Z. Xu}, \bibinfo{person}{Kihyuk Sohn}, \bibinfo{person}{Wenhan Lu}, \bibinfo{person}{David Berthelot}, \bibinfo{person}{Izzeddin Gur}, \bibinfo{person}{Pranav Khaitan}, \bibinfo{person}{Ke Huang}, \bibinfo{person}{Kyriacos~M. Koupparis}, {and} \bibinfo{person}{Bernhard Kowatsch}.} \bibinfo{year}{2020}\natexlab{}.
\newblock \showarticletitle{Assessing Post-Disaster Damage from Satellite Imagery using Semi-Supervised Learning Techniques}.
\newblock \bibinfo{journal}{\emph{ArXiv}}  \bibinfo{volume}{abs/2011.14004} (\bibinfo{year}{2020}).
\newblock
\urldef\tempurl%
\url{https://api.semanticscholar.org/CorpusID:227227614}
\showURL{%
\tempurl}


\bibitem[Lewis and Gale(1994)]%
        {10.5555/188490.188495}
\bibfield{author}{\bibinfo{person}{David~D. Lewis} {and} \bibinfo{person}{William~A. Gale}.} \bibinfo{year}{1994}\natexlab{}.
\newblock \showarticletitle{A Sequential Algorithm for Training Text Classifiers}. In \bibinfo{booktitle}{\emph{Proceedings of the 17th Annual International ACM SIGIR Conference on Research and Development in Information Retrieval}} (Dublin, Ireland) \emph{(\bibinfo{series}{SIGIR '94})}. \bibinfo{publisher}{Springer-Verlag}, \bibinfo{address}{Berlin, Heidelberg}, \bibinfo{pages}{3–12}.
\newblock
\showISBNx{038719889X}


\bibitem[Melville et~al\mbox{.}(2005)]%
        {10.1007/11564096_28}
\bibfield{author}{\bibinfo{person}{Prem Melville}, \bibinfo{person}{Stewart~M. Yang}, \bibinfo{person}{Maytal Saar-Tsechansky}, {and} \bibinfo{person}{Raymond Mooney}.} \bibinfo{year}{2005}\natexlab{}.
\newblock \showarticletitle{Active Learning for Probability Estimation Using Jensen-Shannon Divergence}. In \bibinfo{booktitle}{\emph{Proceedings of the 16th European Conference on Machine Learning}} (Porto, Portugal) \emph{(\bibinfo{series}{ECML'05})}. \bibinfo{publisher}{Springer-Verlag}, \bibinfo{address}{Berlin, Heidelberg}, \bibinfo{pages}{268–279}.
\newblock
\showISBNx{3540292438}
\urldef\tempurl%
\url{https://doi.org/10.1007/11564096_28}
\showDOI{\tempurl}


\bibitem[Nguyen and Smeulders(2004)]%
        {2004Active}
\bibfield{author}{\bibinfo{person}{Hieu~T. Nguyen} {and} \bibinfo{person}{Arnold Smeulders}.} \bibinfo{year}{2004}\natexlab{}.
\newblock \showarticletitle{Active learning using pre-clustering}.
\newblock  (\bibinfo{year}{2004}).
\newblock


\bibitem[Pan and Yang(2010)]%
        {5288526}
\bibfield{author}{\bibinfo{person}{Sinno~Jialin Pan} {and} \bibinfo{person}{Qiang Yang}.} \bibinfo{year}{2010}\natexlab{}.
\newblock \showarticletitle{A Survey on Transfer Learning}.
\newblock \bibinfo{journal}{\emph{IEEE Transactions on Knowledge and Data Engineering}} \bibinfo{volume}{22}, \bibinfo{number}{10} (\bibinfo{year}{2010}), \bibinfo{pages}{1345--1359}.
\newblock
\urldef\tempurl%
\url{https://doi.org/10.1109/TKDE.2009.191}
\showDOI{\tempurl}


\bibitem[Ren et~al\mbox{.}(2021)]%
        {10.1145/3472291}
\bibfield{author}{\bibinfo{person}{Pengzhen Ren}, \bibinfo{person}{Yun Xiao}, \bibinfo{person}{Xiaojun Chang}, \bibinfo{person}{Po-Yao Huang}, \bibinfo{person}{Zhihui Li}, \bibinfo{person}{Brij~B. Gupta}, \bibinfo{person}{Xiaojiang Chen}, {and} \bibinfo{person}{Xin Wang}.} \bibinfo{year}{2021}\natexlab{}.
\newblock \showarticletitle{A Survey of Deep Active Learning}.
\newblock \bibinfo{journal}{\emph{ACM Comput. Surv.}} \bibinfo{volume}{54}, \bibinfo{number}{9}, Article \bibinfo{articleno}{180} (\bibinfo{date}{oct} \bibinfo{year}{2021}), \bibinfo{numpages}{40}~pages.
\newblock
\showISSN{0360-0300}
\urldef\tempurl%
\url{https://doi.org/10.1145/3472291}
\showDOI{\tempurl}


\bibitem[Roth and Small(2006)]%
        {2006Margin}
\bibfield{author}{\bibinfo{person}{Dan Roth} {and} \bibinfo{person}{Kevin Small}.} \bibinfo{year}{2006}\natexlab{}.
\newblock \showarticletitle{Margin-Based Active Learning for Structured Output Spaces}.
\newblock \bibinfo{journal}{\emph{lecture notes in computer science}} (\bibinfo{year}{2006}).
\newblock


\bibitem[Rudner et~al\mbox{.}(2019)]%
        {rudner2019multi3net}
\bibfield{author}{\bibinfo{person}{Tim~GJ Rudner}, \bibinfo{person}{Marc Ru{\ss}wurm}, \bibinfo{person}{Jakub Fil}, \bibinfo{person}{Ramona Pelich}, \bibinfo{person}{Benjamin Bischke}, \bibinfo{person}{Veronika Kopa{\v{c}}kov{\'a}}, {and} \bibinfo{person}{Piotr Bili{\'n}ski}.} \bibinfo{year}{2019}\natexlab{}.
\newblock \showarticletitle{Multi3net: segmenting flooded buildings via fusion of multiresolution, multisensor, and multitemporal satellite imagery}. In \bibinfo{booktitle}{\emph{Proceedings of the AAAI Conference on Artificial Intelligence}}, Vol.~\bibinfo{volume}{33}. \bibinfo{pages}{702--709}.
\newblock


\bibitem[Saar-Tsechansky and Provost(2004)]%
        {10.1023/B:MACH.0000011806.12374.c3}
\bibfield{author}{\bibinfo{person}{Maytal Saar-Tsechansky} {and} \bibinfo{person}{Foster Provost}.} \bibinfo{year}{2004}\natexlab{}.
\newblock \showarticletitle{Active Sampling for Class Probability Estimation and Ranking}.
\newblock \bibinfo{journal}{\emph{Mach. Learn.}} \bibinfo{volume}{54}, \bibinfo{number}{2} (\bibinfo{date}{feb} \bibinfo{year}{2004}), \bibinfo{pages}{153–178}.
\newblock
\showISSN{0885-6125}
\urldef\tempurl%
\url{https://doi.org/10.1023/B:MACH.0000011806.12374.c3}
\showDOI{\tempurl}


\bibitem[Settles(2010)]%
        {2010Active}
\bibfield{author}{\bibinfo{person}{Burr Settles}.} \bibinfo{year}{2010}\natexlab{}.
\newblock \showarticletitle{Active Learning Literature Survey}.
\newblock \bibinfo{journal}{\emph{University of Wisconsinmadison}} (\bibinfo{year}{2010}).
\newblock


\bibitem[Settles and Craven(2008)]%
        {2008An}
\bibfield{author}{\bibinfo{person}{Burr Settles} {and} \bibinfo{person}{Mark Craven}.} \bibinfo{year}{2008}\natexlab{}.
\newblock \showarticletitle{An Analysis of Active Learning Strategies for Sequence Labeling Tasks}. In \bibinfo{booktitle}{\emph{2008 Conference on Empirical Methods in Natural Language Processing, EMNLP 2008, Proceedings of the Conference, 25-27 October 2008, Honolulu, Hawaii, USA, A meeting of SIGDAT, a Special Interest Group of the ACL}}.
\newblock


\bibitem[Shannon(1948)]%
        {1948mathematical}
\bibfield{author}{\bibinfo{person}{C~E Shannon, A}.} \bibinfo{year}{1948}\natexlab{}.
\newblock \bibinfo{booktitle}{\emph{mathematical theory of communication}}.
\newblock \bibinfo{publisher}{mathematical theory of communication}.
\newblock


\bibitem[Simonyan and Zisserman(2014)]%
        {2014Very}
\bibfield{author}{\bibinfo{person}{Karen Simonyan} {and} \bibinfo{person}{Andrew Zisserman}.} \bibinfo{year}{2014}\natexlab{}.
\newblock \showarticletitle{Very Deep Convolutional Networks for Large-Scale Image Recognition}.
\newblock \bibinfo{journal}{\emph{Computer Science}} (\bibinfo{year}{2014}).
\newblock


\bibitem[Tang and Huang(2019)]%
        {Tang_Huang_2019}
\bibfield{author}{\bibinfo{person}{Ying-Peng Tang} {and} \bibinfo{person}{Sheng-Jun Huang}.} \bibinfo{year}{2019}\natexlab{}.
\newblock \showarticletitle{Self-Paced Active Learning: Query the Right Thing at the Right Time}.
\newblock \bibinfo{journal}{\emph{Proceedings of the AAAI Conference on Artificial Intelligence}} \bibinfo{volume}{33}, \bibinfo{number}{01} (\bibinfo{date}{Jul.} \bibinfo{year}{2019}), \bibinfo{pages}{5117--5124}.
\newblock
\urldef\tempurl%
\url{https://doi.org/10.1609/aaai.v33i01.33015117}
\showDOI{\tempurl}


\bibitem[Van~Etten et~al\mbox{.}(2018)]%
        {van2018spacenet}
\bibfield{author}{\bibinfo{person}{Adam Van~Etten}, \bibinfo{person}{Dave Lindenbaum}, {and} \bibinfo{person}{Todd~M Bacastow}.} \bibinfo{year}{2018}\natexlab{}.
\newblock \showarticletitle{Spacenet: A remote sensing dataset and challenge series}.
\newblock \bibinfo{journal}{\emph{arXiv preprint arXiv:1807.01232}} (\bibinfo{year}{2018}).
\newblock


\bibitem[Wang and Shang(2014)]%
        {6889457}
\bibfield{author}{\bibinfo{person}{Dan Wang} {and} \bibinfo{person}{Yi Shang}.} \bibinfo{year}{2014}\natexlab{}.
\newblock \showarticletitle{A new active labeling method for deep learning}. In \bibinfo{booktitle}{\emph{2014 International Joint Conference on Neural Networks (IJCNN)}}. \bibinfo{pages}{112--119}.
\newblock
\urldef\tempurl%
\url{https://doi.org/10.1109/IJCNN.2014.6889457}
\showDOI{\tempurl}


\bibitem[Xia et~al\mbox{.}(2022)]%
        {xia2022self}
\bibfield{author}{\bibinfo{person}{Zaishuo Xia}, \bibinfo{person}{Zelin Li}, \bibinfo{person}{Yanbing Bai}, \bibinfo{person}{Jinze Yu}, {and} \bibinfo{person}{Bruno Adriano}.} \bibinfo{year}{2022}\natexlab{}.
\newblock \showarticletitle{Self-supervised learning for building damage assessment from large-scale xBD satellite imagery benchmark datasets}. In \bibinfo{booktitle}{\emph{International Conference on Database and Expert Systems Applications}}. Springer, \bibinfo{pages}{373--386}.
\newblock


\bibitem[Xie et~al\mbox{.}(2021)]%
        {DBLP:journals/corr/abs-2112-01406}
\bibfield{author}{\bibinfo{person}{Binhui Xie}, \bibinfo{person}{Longhui Yuan}, \bibinfo{person}{Shuang Li}, \bibinfo{person}{Chi~Harold Liu}, \bibinfo{person}{Xinjing Cheng}, {and} \bibinfo{person}{Guoren Wang}.} \bibinfo{year}{2021}\natexlab{}.
\newblock \showarticletitle{Active Learning for Domain Adaptation: An Energy-based Approach}.
\newblock \bibinfo{journal}{\emph{CoRR}}  \bibinfo{volume}{abs/2112.01406} (\bibinfo{year}{2021}).
\newblock
\showeprint[arXiv]{2112.01406}
\urldef\tempurl%
\url{https://arxiv.org/abs/2112.01406}
\showURL{%
\tempurl}


\bibitem[Xu et~al\mbox{.}(2019a)]%
        {xu2019building}
\bibfield{author}{\bibinfo{person}{Joseph~Z Xu}, \bibinfo{person}{Wenhan Lu}, \bibinfo{person}{Zebo Li}, \bibinfo{person}{Pranav Khaitan}, {and} \bibinfo{person}{Valeriya Zaytseva}.} \bibinfo{year}{2019}\natexlab{a}.
\newblock \showarticletitle{Building damage detection in satellite imagery using convolutional neural networks}.
\newblock \bibinfo{journal}{\emph{arXiv preprint arXiv:1910.06444}} (\bibinfo{year}{2019}).
\newblock


\bibitem[Xu et~al\mbox{.}(2019b)]%
        {2019Building}
\bibfield{author}{\bibinfo{person}{Joseph~Z Xu}, \bibinfo{person}{Wenhan Lu}, \bibinfo{person}{Zebo Li}, \bibinfo{person}{Pranav Khaitan}, {and} \bibinfo{person}{Valeriya Zaytseva}.} \bibinfo{year}{2019}\natexlab{b}.
\newblock \showarticletitle{Building Damage Detection in Satellite Imagery Using Convolutional Neural Networks}.
\newblock \bibinfo{journal}{\emph{arXiv}} (\bibinfo{year}{2019}).
\newblock


\bibitem[Yang et~al\mbox{.}(2015)]%
        {10.1007/s11263-014-0781-x}
\bibfield{author}{\bibinfo{person}{Yi Yang}, \bibinfo{person}{Zhigang Ma}, \bibinfo{person}{Feiping Nie}, \bibinfo{person}{Xiaojun Chang}, {and} \bibinfo{person}{Alexander~G. Hauptmann}.} \bibinfo{year}{2015}\natexlab{}.
\newblock \showarticletitle{Multi-Class Active Learning by Uncertainty Sampling with Diversity Maximization}.
\newblock \bibinfo{journal}{\emph{Int. J. Comput. Vision}} \bibinfo{volume}{113}, \bibinfo{number}{2} (\bibinfo{date}{jun} \bibinfo{year}{2015}), \bibinfo{pages}{113–127}.
\newblock
\showISSN{0920-5691}
\urldef\tempurl%
\url{https://doi.org/10.1007/s11263-014-0781-x}
\showDOI{\tempurl}


\bibitem[Yoo and Kweon(2019)]%
        {8954021}
\bibfield{author}{\bibinfo{person}{Donggeun Yoo} {and} \bibinfo{person}{In~So Kweon}.} \bibinfo{year}{2019}\natexlab{}.
\newblock \showarticletitle{Learning Loss for Active Learning}. In \bibinfo{booktitle}{\emph{2019 IEEE/CVF Conference on Computer Vision and Pattern Recognition (CVPR)}}. \bibinfo{pages}{93--102}.
\newblock
\urldef\tempurl%
\url{https://doi.org/10.1109/CVPR.2019.00018}
\showDOI{\tempurl}


\bibitem[Zhang and Chen(2002)]%
        {1017738}
\bibfield{author}{\bibinfo{person}{Cha Zhang} {and} \bibinfo{person}{Tsuhan Chen}.} \bibinfo{year}{2002}\natexlab{}.
\newblock \showarticletitle{An active learning framework for content-based information retrieval}.
\newblock \bibinfo{journal}{\emph{IEEE Transactions on Multimedia}} \bibinfo{volume}{4}, \bibinfo{number}{2} (\bibinfo{year}{2002}), \bibinfo{pages}{260--268}.
\newblock
\urldef\tempurl%
\url{https://doi.org/10.1109/TMM.2002.1017738}
\showDOI{\tempurl}


\bibitem[Zhang et~al\mbox{.}(2022)]%
        {2022BoostMIS}
\bibfield{author}{\bibinfo{person}{Wenqiao Zhang}, \bibinfo{person}{Lei Zhu}, \bibinfo{person}{James Hallinan}, \bibinfo{person}{Andrew Makmur}, \bibinfo{person}{Shengyu Zhang}, \bibinfo{person}{Qingpeng Cai}, {and} \bibinfo{person}{Beng~Chin Ooi}.} \bibinfo{year}{2022}\natexlab{}.
\newblock \showarticletitle{BoostMIS: Boosting Medical Image Semi-supervised Learning with Adaptive Pseudo Labeling and Informative Active Annotation}.
\newblock \bibinfo{journal}{\emph{arXiv e-prints}} (\bibinfo{year}{2022}).
\newblock


\bibitem[Zhou et~al\mbox{.}(2017)]%
        {8099989}
\bibfield{author}{\bibinfo{person}{Zongwei Zhou}, \bibinfo{person}{Jae Shin}, \bibinfo{person}{Lei Zhang}, \bibinfo{person}{Suryakanth Gurudu}, \bibinfo{person}{Michael Gotway}, {and} \bibinfo{person}{Jianming Liang}.} \bibinfo{year}{2017}\natexlab{}.
\newblock \showarticletitle{Fine-Tuning Convolutional Neural Networks for Biomedical Image Analysis: Actively and Incrementally}. In \bibinfo{booktitle}{\emph{2017 IEEE Conference on Computer Vision and Pattern Recognition (CVPR)}}. \bibinfo{pages}{4761--4772}.
\newblock
\urldef\tempurl%
\url{https://doi.org/10.1109/CVPR.2017.506}
\showDOI{\tempurl}


\end{thebibliography}




\nocite{*} 
\end{document}